\documentclass{article}
\PassOptionsToPackage{numbers}{natbib}
\usepackage[final]{neurips_2023}
\makeatletter
\renewcommand{\@noticestring}{}
\makeatother

\usepackage[utf8]{inputenc}
\usepackage[T1]{fontenc}
\usepackage{hyperref}
\usepackage{url}
\usepackage{booktabs}
\usepackage{amsfonts}
\usepackage{amsmath}
\usepackage{nicefrac}
\usepackage{microtype}
\usepackage{xcolor}
\usepackage{graphicx}
\usepackage{algorithm}
\usepackage{algorithmic}
\usepackage{enumitem}
\usepackage{tikz}
\usetikzlibrary{arrows.meta,positioning,shapes.geometric,calc,decorations.pathreplacing}
\usepackage{pgfplots}
\pgfplotsset{compat=1.18}
\usepackage{listings}
\usepackage{multirow}
\usepackage{pifont}

\newcommand{\cmark}{\ding{51}}
\newcommand{\xmark}{\ding{55}}
\newcommand{\pmark}{\textbf{$\sim$}}

\title{AutoMegaKernel: A Statically-Checked Agent Harness for Self-Retargeting Megakernel Synthesis}

\author{%
  Jaber Jaber\thanks{Correspondence: \texttt{jaber@rightnowai.co}} \\
  RightNow AI \\
  \texttt{jaber@rightnowai.co} \\
  \And
  Osama Jaber \\
  RightNow AI \\
  \texttt{osama@rightnowai.co} \\
}

\begin{document}

\maketitle

\begin{center}
\includegraphics[height=1.1cm]{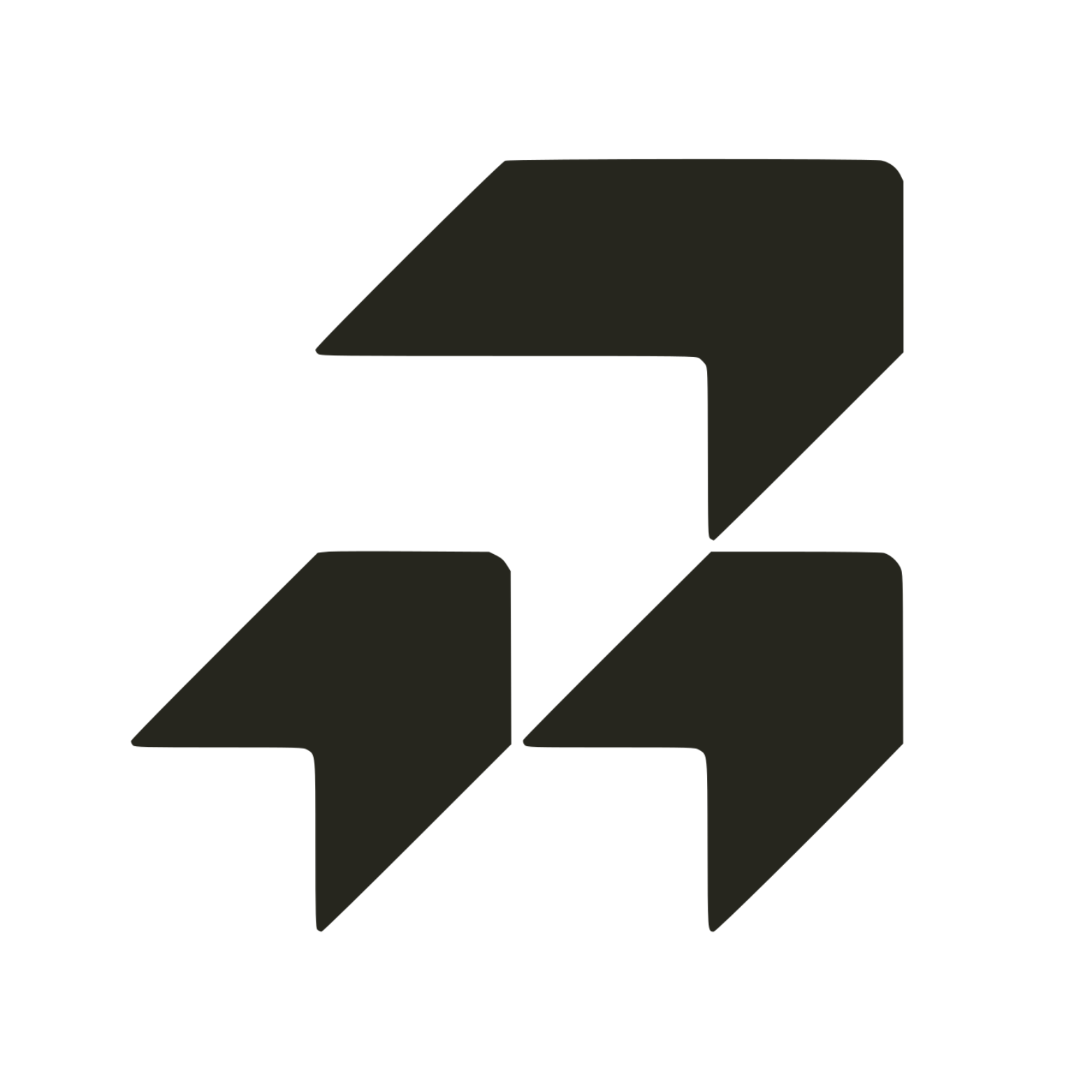}
\end{center}

\begin{abstract}
Single-stream LLM decode is bandwidth-bound: each token streams the whole weight set through the SMs once, so the latency floor is $\text{weights}/\text{HBM bandwidth}$. Standard execution launches one kernel per operator and round-trips activations through HBM between every op. \textsc{AutoMegaKernel} (\textsc{AMK}) compiles a HuggingFace Llama-family model into a single persistent cooperative kernel that runs the entire forward pass in one launch, with no per-model hand-written CUDA. The contribution is the system, not raw speed.

A frozen schedule-IR validator statically certifies deadlock-freedom (DAG acyclicity, wait-satisfiability $1 \le t \le \#\text{producers}$, per-SM queue order) and race-freedom (shared-counter all-join rule, transitive happens-before provenance) via a set of static graph checks (not a mechanized proof), so an unsafe agent-proposed schedule is rejected before launch. Across 7{,}160 adversarially-constructed schedules (6{,}091 unsafe), the validator had zero false-accepts and accepted all 360 real lowerings, 24 of which matched eager bit-for-bit. The same source retargets sm\_80, sm\_90, and sm\_120 from one codebase, auto-generates correct megakernels for 10 of 10 supported HF models, and on a real SmolLM2-135M checkpoint reproduces HuggingFace greedy decode token-for-token and matches teacher-forced perplexity to $2.5\times10^{-7}$.

\textsc{AMK} also auto-generates int8 and int4 weight-only quantized megakernels: int8 is greedy-lossless and $1.12\times$ faster per token; int4 cuts the weight-traffic floor $2.42\times$ (blended) at a documented accuracy cost. An unattended, agent-drivable autoresearch loop autonomously improves the generated megakernel over its own baseline ($1.25$--$1.72\times$ measured).

The performance study reports every direction. The strongest real-hardware result is an \emph{inference-class} cuBLAS win: a search-found int8 weight-only (W8A16) megakernel beats CUDA-graphed cuBLAS bf16 at batch-1 decode across the datacenter inference fleet: the \textbf{NVIDIA L4} by up to $1.33\times$ (growing with model size), the current-gen \textbf{L40S} by $1.25$--$1.27\times$, and the \textbf{A10G} by up to $1.08\times$ at scale, and on the consumer RTX 5090 Laptop by $1.19$--$1.23\times$, a precision-asymmetric comparison (W8A16 vs.\ bf16), measured on random-initialized weights at real Llama shapes (batch-1 latency is shape/byte-determined; the near-lossless W8A16 \emph{quality} is shown on the real SmolLM2-135M checkpoint). The ordering is not a clean function of bandwidth (the $864$\,GB/s L40S wins by more than the $600$\,GB/s A10G); the dividing line is the inference-class vs.\ training-class regime. It trails cuBLAS on the training-class A100/H100, where the harness localizes the cross-SM-sync bottleneck, the honest boundary, which we disclose rather than hide. All latencies are batch-1 at decode position~0 (empty KV); the largest real checkpoint is TinyLlama-1.1B. The optimized bf16 GEMV reaches ${\approx}460$\,GB/s ($63\%$ of measured peak) versus a cuBLAS ceiling of ${\approx}90\%$; we report the gap plainly. Code, data, and the agent harness are open-source at \url{https://github.com/RightNow-AI/AutoMegaKernel}.\thanks{\url{https://www.rightnowai.co/}}
\end{abstract}

\section{Introduction}

A single decode step of an autoregressive language model at batch 1 reads every weight once to produce one token. The arithmetic intensity is near 1, so the step is bound by memory bandwidth: its floor is $t_{\min} = \text{weight\_bytes} / \text{HBM\_bandwidth}$~\cite{williams2009roofline}. A conventional PyTorch or cuBLAS~\cite{nvidia_cublas} execution does not approach that floor. It launches one kernel per operator, pays CPU launch latency dozens of times per layer, and round-trips activations through HBM at every op boundary~\cite{ansel2024pytorch2}. CUDA Graphs~\cite{gray2019cudagraphs} amortize launch overhead by replaying a captured op sequence, but the inter-op kernel boundaries and their HBM round-trips remain.

A megakernel removes those boundaries. It launches once, keeps a persistent threadblock resident on every SM, and walks the model's dependency graph in place. Recent work shows this is the right shape for low-latency decode: Mirage Persistent Kernel (MPK) auto-transforms a tensor program into a single persistent kernel with in-kernel scheduling~\cite{cheng2025mpk}, and a hand-built Llama-1B megakernel removes all launch bubbles, observing that vLLM and SGLang use at most half of H100 bandwidth at low latency~\cite{spector2025nobubbles, kwon2023pagedattention, zheng2023sglang}. The gap these systems leave open is trust and portability. MPK ships no static deadlock/race gate; the hand-built kernel targets one model on one architecture; neither is designed as an edit surface a coding agent can drive safely.

\begin{figure}[t]
\centering
\definecolor{amkblue}{RGB}{31,86,140}
\definecolor{amkgreen}{RGB}{34,123,75}
\definecolor{amkgray}{RGB}{95,95,95}
\resizebox{\linewidth}{!}{%
\begin{tikzpicture}[
  >={Stealth[length=2.0mm]},
  font=\small,
  stage/.style={draw=amkblue!80, line width=0.5pt, rounded corners=2.5pt,
                fill=amkblue!7, align=center, inner xsep=4pt, inner ysep=3pt,
                font=\small, text width=2.05cm, minimum height=11mm},
  gate/.style={draw=amkgreen!85, line width=0.8pt, rounded corners=2.5pt,
               fill=amkgreen!10, align=center, inner xsep=4pt, inner ysep=3pt,
               font=\small, text width=2.05cm, minimum height=11mm},
  term/.style={draw=amkgray, line width=0.5pt, rounded corners=2.5pt,
               fill=amkgray!8, align=center, inner sep=4pt, font=\small,
               text width=1.55cm, minimum height=11mm},
  kern/.style={draw=amkblue!85, line width=1.0pt, rounded corners=3pt,
               fill=amkblue!12, align=center, inner sep=4pt, font=\small,
               text width=4.7cm, minimum height=12mm},
  sm/.style={draw=amkblue!70, line width=0.4pt, rounded corners=1.5pt,
             fill=amkblue!6, align=center, inner sep=1.5pt, font=\scriptsize,
             minimum width=12mm, minimum height=8mm},
  flow/.style={->, amkblue!85, line width=0.9pt},
  note/.style={font=\scriptsize\itshape, amkgray, align=center}
]
\node[stage] (hf) at (0,0) {\textbf{HF Llama}\\ model};
\node[stage, right=6mm of hf] (ir)
  {\textbf{Schedule IR}\\[1pt] {\scriptsize SM-level task-DAG, counter-sync'd}};
\node[gate, right=6mm of ir] (val)
  {\texttt{validate()}\\[1pt] {\scriptsize static deadlock + race-free gate}};
\node[note, above=0.8mm of val] {0 false-accepts / 7160 schedules};

\node[kern, below=13mm of hf.south west, anchor=north west] (mk)
  {\textbf{Persistent cooperative megakernel}\\[1pt]
   {\scriptsize one threadblock per SM, \texttt{grid.sync}, counter producer/consumer}};
\node[term, right=7mm of mk] (tok) {\textbf{decoded token}};
\node[note, below=0.8mm of mk.south] {1 launch $=$ 1 forward pass $=$ 1 token};

\draw[flow] (hf) -- (ir);
\draw[flow] (ir) -- (val);
\draw[flow] (val.south) -- ++(0,-3mm) -| (mk.north);
\draw[flow] (mk) -- (tok);

\node[draw=amkgray!55, dashed, rounded corners=2pt, inner sep=3pt,
      font=\scriptsize\itshape, amkgray, align=center, right=8mm of val]
  (rt) {retargets\\ sm\_80 / sm\_90 / sm\_120\\ from one source};

\node[sm, below=14mm of tok.north, anchor=north] (smB) {SM$_1$\\ wait $\ge t$};
\node[sm, left=2.5mm of smB] (smA) {SM$_0$\\ ctr\,\texttt{++}};
\node[sm, right=2.5mm of smB] (smC) {SM$_2$\\ ctr\,\texttt{++}};
\draw[->, amkgreen!85, line width=0.7pt] (smA.east) -- (smB.west);
\draw[->, amkgreen!85, line width=0.7pt] (smB.east) -- (smC.west);
\node[note, text width=4.2cm, below=0.6mm of smB.south]
  {per-SM counters: producer \texttt{++}, consumer waits on threshold};
\end{tikzpicture}%
}
\caption{\textbf{Correctness-by-construction compilation pipeline.}
A HuggingFace Llama model lowers to a typed \emph{schedule IR}, an SM-level
task-DAG whose only cross-task signalling is monotonic counters, then passes
the static \texttt{validate()} gate, which certifies deadlock- and
race-freedom \emph{before} any launch (0 false-accepts over 7{,}160
adversarial schedules). The accepted schedule runs as one \emph{persistent
cooperative megakernel}: a single \texttt{cudaLaunchCooperativeKernel}
co-resides one threadblock per SM, synchronizing with \texttt{grid.sync} and
counter-based producer/consumer handoffs (inset), so one launch is one forward
pass is one decoded token. The same source retargets sm\_80/sm\_90/sm\_120.}
\label{fig:pipeline}
\end{figure}
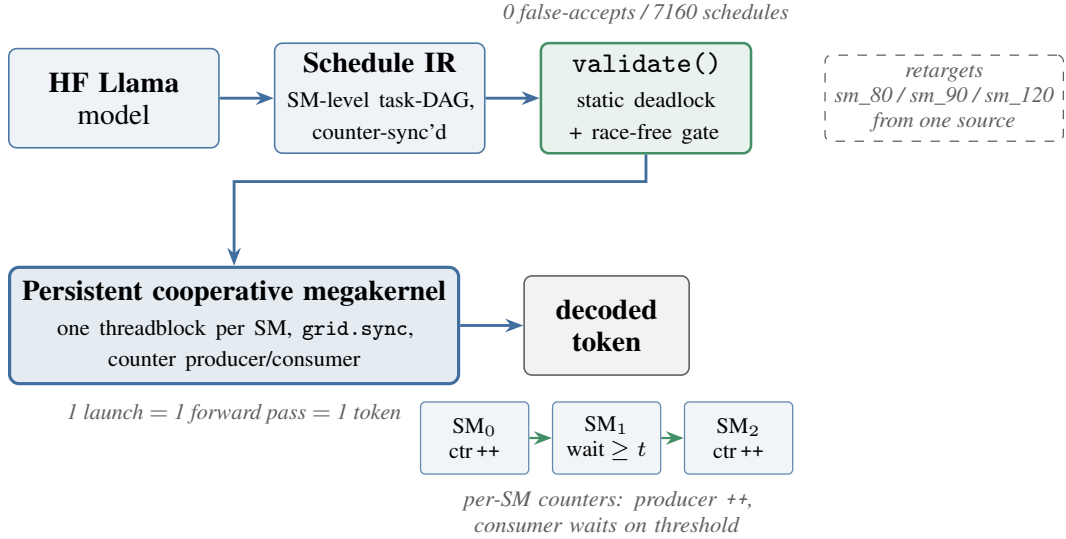

\textsc{AMK} extends \textsc{AutoKernel}~\cite{jaber2026autokernel}, our autonomous, agent-driven GPU kernel optimizer, which achieved remarkable single-kernel results: up to $5.29\times$ over PyTorch eager via iterative agent-driven search. \textsc{AMK} carries that same propose~$\to$~evaluate~$\to$~keep/revert search methodology up one level of abstraction: from optimizing one kernel at a time to compiling and self-optimizing an \emph{entire model} as a single persistent megakernel. \textsc{AMK} treats correctness as a structural invariant of the compiler, not a property of the output. The forward pass lowers to a typed schedule IR over an SM-level task-DAG synchronized by monotonic counters, run by a four-layer system (Section~\ref{sec:design}; Figure~\ref{fig:pipeline} shows the pipeline). A validator statically checks the schedule is deadlock-free and race-free before any launch (a set of graph checks over a trusted hand-written base, not a mechanized proof). The key insight is that a forward pass is a DAG, producers only increment counters, and consumers only wait on statically known thresholds, so safety reduces to a small set of static graph checks that an automated agent cannot violate at runtime. An invalid schedule becomes a clean \texttt{REJECTED} at validation time instead of a hung GPU.

Our contributions:
\begin{enumerate}[leftmargin=1.4em,itemsep=2pt]
  \item \textbf{A statically-checked schedule IR, stress-tested.} A frozen validator (\texttt{schedule/ir.py}, 1150 lines) statically \emph{checks} deadlock-freedom and race-freedom: a set of static graph checks, not a mechanized proof, over a trusted hand-written base (the validator itself and the per-architecture VM). The on-device VM refuses to load anything it rejects (Section~\ref{sec:tech}). Across 7{,}160 adversarial schedules (6{,}091 unsafe by an independent oracle, spanning 8 unsafe classes), the validator had \emph{zero false-accepts} and accepted all 360 real lowerings, 24 re-run bit-for-bit vs.\ eager (Section~\ref{sec:eval}). This is empirical soundness over a trusted base, not formal verification; but it is a static safety gate that MPK and hand-built megakernels do not provide.
  \item \textbf{Automatic generation coverage.} \textsc{AMK} auto-generates a correct megakernel for 10 of 10 supported models with zero hand-written CUDA: three real HF checkpoints (SmolLM2-135M/360M, TinyLlama-1.1B, up to 3{,}410 IR tasks) plus a 40M--618M from-config sweep, all token-for-token vs eager, and rejects 3 of 4 unsupported variants loudly (the fourth gap documented honestly) (Section~\ref{sec:eval}).
  \item \textbf{Self-retargeting.} The same source built and ran a correct megakernel on sm\_120 (RTX 5090), sm\_80 (A100), and sm\_90 (H100), gencode auto-derived from the live device, matching eager to $\le 4.2\times10^{-7}$ (fp32) on the toy and synthetic models and $3.8\times10^{-5}$ on the real SmolLM2-135M checkpoint, well within the $10^{-4}$ fp32 tolerance (Table~\ref{tab:correctness}).
  \item \textbf{An agent-drivable compiler.} The Layer-2 edit surface is a structured \texttt{ScheduleConfig}, not kernel code; the frozen VM deterministically lowers it and the validator gates it (Section~\ref{sec:tech}).
  \item \textbf{A real-checkpoint path with quantization.} \textsc{AMK} imports SmolLM2-135M, reproduces HuggingFace greedy decode token-for-token, and matches HuggingFace teacher-forced perplexity to $2.5\times10^{-7}$ (Section~\ref{sec:eval}). It also auto-generates int8 and int4 weight-only quantized megakernels: int8 is greedy-lossless and $1.12\times$ faster per token; int4 is correct-but-lossy and cuts the weight-traffic floor $2.42\times$.
  \item \textbf{A self-improving, agent-drivable harness.} Beyond a static edit surface, \textsc{AMK} ships an unattended knob-autotuning loop (propose~$\to$~\texttt{validate}~$\to$~correctness-gate~$\to$~measure~$\to$~keep/revert) over a small schedule-and-kernel-knob grid (cols/warp, tile width, threads, \texttt{cp.async} depth), with drift-robust per-sample-interleaved measurement and a physical-roofline-floor honesty guard. It improves the megakernel \emph{over its own default} ($1.25$--$1.72\times$, a self-relative gain, not a win over an external baseline) (Section~\ref{sec:eval}).
  \item \textbf{An honest performance study, with a real cuBLAS win on inference-class GPUs.} Wall-clock CUDA-event timing plus an analytic roofline against both spec and measured HBM peak, across seven GPUs (Section~\ref{sec:eval}). The auto-tuned int8 weight-only megakernel \emph{outperforms} CUDA-graphed cuBLAS bf16 at batch-1 across the datacenter \emph{inference-class} fleet: the NVIDIA L4 by up to $1.33\times$ (growing with model size), the current-gen L40S by $1.25$--$1.27\times$, and the A10G by up to $1.08\times$ at scale, and on the consumer RTX 5090 by $1.19$--$1.23\times$. The ordering is not a clean function of bandwidth (the $864$\,GB/s L40S wins by more than the $600$\,GB/s A10G), so the dividing line is the inference-class vs.\ training-class regime; this is a precision-asymmetric comparison (W8A16 vs.\ bf16), and we say so. It trails cuBLAS on the training-class A100/H100, where the harness localizes the cross-SM-sync bottleneck, the honest boundary, disclosed in the Limitations (Section~\ref{sec:limitations}). All latencies are batch-1 at decode position~0; the largest real checkpoint is TinyLlama-1.1B. Figure~\ref{fig:l4-selfimprove} summarizes; we report every direction plainly.
\end{enumerate}

\section{Related Work}
\label{sec:related}

\paragraph{Megakernels and persistent execution.} MPK~\cite{cheng2025mpk} is the closest system: a compiler and runtime that mega-kernelizes a tensor program through an SM-level task graph with decentralized in-kernel scheduling, cutting end-to-end latency up to $1.7\times$. \textsc{AMK} targets the same single-launch shape but adds a static safety gate and a structured agent edit surface. The hand-built Llama-1B megakernel of Spector et al.~\cite{spector2025nobubbles} removes launch bubbles for one model; \textsc{AMK} auto-generates the megakernel from any HF Llama-family model with no per-model CUDA, at the cost of a kernel quality gap we report.

\paragraph{Serving systems.} vLLM~\cite{kwon2023pagedattention}, SGLang~\cite{zheng2023sglang}, Orca~\cite{yu2022orca}, and TensorRT-LLM~\cite{nvidia2023tensorrtllm} optimize high-batch throughput with per-op or per-engine kernels and advanced batching and KV management. \textsc{AMK} optimizes single-stream batch-1 latency and does not beat these systems on throughput, nor does it claim to.

\paragraph{Kernel authoring and attention.} FlashAttention and its successors~\cite{dao2022flashattention, dao2023flashattention2, shah2024flashattention3} and Flash-Decoding~\cite{dao2023flashdecoding} optimize attention as a best-in-class op; Triton~\cite{tillet2019triton} and ThunderKittens~\cite{spector2024thunderkittens} are kernel-authoring frameworks. These are instruction-level optimizations \textsc{AMK} consumes as Layer-1 micro-kernels rather than alternatives to whole-model fusion.

\paragraph{Compilers and auto-schedulers.} TVM~\cite{chen2018tvm}, AutoTVM~\cite{chen2018autotvm}, Ansor~\cite{zheng2020ansor}, Hidet~\cite{ding2023hidet}, Welder~\cite{shi2023welder}, and PyTorch~2~\cite{ansel2024pytorch2} schedule operators or fused groups, but emit a graph of separate kernel launches. \textsc{AMK}'s search axis is the whole-model schedule realized by one persistent kernel, gated by a static correctness certificate.

\paragraph{Weight-only quantized GEMV.} Hand-built mixed-precision kernels such as Marlin~\cite{frantar2024marlin} reach near-roofline batch-1 throughput for weight-only-quantized (e.g.\ int4/int8 W$n$A16) GEMV by carefully engineering memory-level parallelism and dequant scheduling for one precision and layout. \textsc{AMK} instead \emph{auto-generates} its int8/int4 weight-only GEMV from the same schedule-IR path with no per-precision hand CUDA, folding dequant into the GEMV; the trade-off is a kernel-quality gap to such hand-built kernels (our int8 win over cuBLAS bf16 on the RTX 5090 is precision-asymmetric, from streaming fewer bytes, not a faster per-byte kernel), and a Marlin-class quantized GEMV is exactly the Layer-1 micro-kernel \textsc{AMK} would consume to close it.

\paragraph{Orthogonal decode acceleration.} Speculative decoding and its variants~\cite{leviathan2023speculative, chen2023speculativesampling, cai2024medusa, li2024eagle} cut the number of serial steps. \textsc{AMK} cuts the cost of each step's execution; the two compose.

Table~\ref{tab:related} compares systems on the five properties \textsc{AMK} claims. \textsc{AMK}'s auto-generated column is partial: schedules are auto-lowered and there is zero per-model hand CUDA, but the Layer-0 VM is a hand-written, frozen, per-arch trusted base.

\begin{table}[t]
\centering
\caption{Property comparison. \cmark{} yes, \xmark{} no, \pmark{} partial. Column meanings in Section~\ref{sec:related}.}
\label{tab:related}
\small
\begin{tabular}{lccccc}
\toprule
System & Whole-model & Auto-gen, & Correctness- & Self- & Agent- \\
       & fused       & no hand CUDA & by-construction & retargeting & drivable \\
\midrule
MPK~\cite{cheng2025mpk}            & \cmark & \cmark & \xmark & \pmark & \xmark \\
vLLM~\cite{kwon2023pagedattention} & \xmark & \xmark & \xmark & \pmark & \xmark \\
SGLang~\cite{zheng2023sglang}      & \xmark & \xmark & \xmark & \pmark & \xmark \\
TensorRT-LLM~\cite{nvidia2023tensorrtllm} & \pmark & \xmark & \xmark & \xmark & \xmark \\
Ansor/TVM~\cite{zheng2020ansor, chen2018tvm} & \xmark & \cmark & \xmark & \cmark & \xmark \\
\textbf{\textsc{AMK}}              & \cmark & \pmark & \cmark & \cmark & \cmark \\
\bottomrule
\end{tabular}
\end{table}

\section{System Design}
\label{sec:design}

\begin{figure}[t]
\centering
\begin{tikzpicture}[
  node distance=5mm,
  layer/.style={draw, rounded corners, align=center, text width=5.4cm, minimum height=8.5mm, font=\small},
  loop/.style={draw, dashed, rounded corners, align=center, text width=2.3cm, font=\scriptsize\itshape},
  >={Stealth[length=2mm]}
]
\node[layer] (l3) {\textbf{Layer 3: Dynamism} (roadmap)\\ continuous batching, dynamic shapes, MoE};
\node[layer, below=of l3] (l2) {\textbf{Layer 2: Scheduler} (\texttt{schedule/})\\ HF model $\to$ graph IR $\to$ task-DAG\\ + \texttt{validate()} safety gate};
\node[layer, below=of l2] (l1) {\textbf{Layer 1: Instructions} (\texttt{instructions/})\\ ABI-conformant micro-kernels\\ (GEMV, attention, RMSNorm, RoPE\dots)};
\node[layer, below=of l1] (l0) {\textbf{Layer 0: VM} (\texttt{vm/})\\ persistent cooperative megakernel\\ per-SM scheduler, counters, pages};
\draw[->] (l3) -- (l2);
\draw[->] (l2) -- (l1);
\draw[->] (l1) -- (l0);

\node[loop, right=6mm of l2] (loop2) {Loop 2:\\ schedule search};
\node[loop, right=6mm of l1] (loop1) {Loop 1:\\ instruction tuning};
\draw[->] (loop2.north) to[out=120,in=20] (l2.east);
\draw[->] (l2.east) to[out=-20,in=-120] (loop2.south);
\draw[->] (loop1.north) to[out=120,in=20] (l1.east);
\draw[->] (l1.east) to[out=-20,in=-120] (loop1.south);
\end{tikzpicture}
\caption{The four layers and two loops. Layer 0 is the trusted, frozen, hand-written per-arch base; Layers 1--2 are searched by agents but gated by isolated correctness checks (Loop 1) and the static schedule validator (Loop 2).}
\label{fig:arch}
\end{figure}
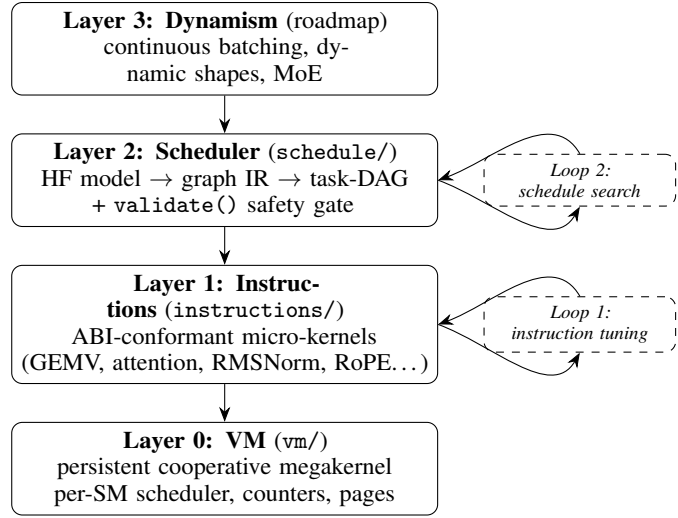

\textsc{AMK} is four layers and two autoresearch loops (Figure~\ref{fig:arch}). Generation is confined inside a verified structure: correctness is a property of the architecture, not of the generated schedule.

\textbf{Layer 0, the VM} (\texttt{vm/scheduler.cu}, 132 lines; \texttt{vm/loader.py}, 857 lines) is the persistent megakernel. One \texttt{cudaLaunchCooperativeKernel} co-resides one block per SM; each block owns one SM-queue and walks it in global topological order, executing \emph{wait} $\to$ \emph{dispatch} $\to$ \emph{signal} per instruction. It is hand-written, exhaustively verified, and frozen per architecture. \textbf{Layer 1} (\texttt{instructions/}) holds ABI-conformant micro-kernels, each correctness-checked against its reference op in isolation. \textbf{Layer 2} (\texttt{schedule/ir.py}, 1150 lines; \texttt{schedule/graph.py}, 548 lines) imports an HF model, lowers it to a tiled task-DAG, and validates it. \textbf{Layer 3} is a roadmap placeholder. Loop 1 edits one micro-kernel under an isolated test; Loop 2 edits a \texttt{ScheduleConfig} that the frozen VM lowers deterministically and the validator gates.

Algorithm~\ref{alg:sm} is the per-SM scheduler loop run by every block (\texttt{vm/scheduler.cu}). Two grid-wide barriers bracket the walk so host-zeroed counters are visible before any spin and all stores are visible after.

\begin{algorithm}[t]
\caption{Per-SM scheduler loop (one block per SM, run cooperatively)}
\label{alg:sm}
\begin{algorithmic}[1]
\STATE \texttt{grid.sync()} \COMMENT{entry barrier: host-zeroed counters + table copies visible}
\IF{$sm < \texttt{num\_sms}$}
  \FOR{each instruction $i$ in SM-queue[$sm$], in global topological order}
    \STATE \textbf{prefetch} next GEMV weight tile to L2 if \texttt{pipelining\_depth}~$>0$ \COMMENT{pure hint}
    \IF{\textbf{not} \texttt{wait\_all}($i$)} \STATE \textbf{break} \COMMENT{abort\_flag set: WDDM/TDR escape} \ENDIF
    \STATE \texttt{dispatch}($i$) \COMMENT{pure block-cooperative compute (Layer 1)}
    \STATE \texttt{signal}($i.\texttt{out\_counter}$) \COMMENT{release fence, then $\texttt{atomicAdd}(+1)$}
  \ENDFOR
\ENDIF
\STATE \texttt{grid.sync()} \COMMENT{exit barrier}
\end{algorithmic}
\end{algorithm}

\section{Technical Details}
\label{sec:tech}

\paragraph{Synchronization model.} Cross-task signalling is only through monotonic \texttt{uint32} counters. Each task, on completion, issues a device-scope release fence ordering all its output-buffer stores, then increments exactly one \texttt{out\_counter} by 1: ``all my outputs are written and visible.'' Before executing, a task waits on a set of \texttt{(counter, threshold)} pairs with statically known thresholds via an acquire-load spin the compiler may not hoist, with exponential backoff and an \texttt{abort\_flag} poll for the WDDM watchdog escape. There are no locks and no arbitrary signalling. One kernel launch is one forward pass is one decoded token; counters are host-memset to zero before each launch and the KV cache persists in HBM across launches.

\paragraph{The invariants \texttt{validate()} proves.} \texttt{schedule.ir.validate} returns a \texttt{ValidationResult} and never raises; a rejected result must prevent launch. Table~\ref{tab:checks} lists the checks. Deadlock-freedom rests on three: every wait threshold satisfies $1 \le t \le \#\text{producers}$ (a wait on a counter with no producer, or above its producer count, is unsatisfiable); the producer$\to$consumer graph is acyclic (checked by Kahn's algorithm with an iterative-DFS cycle witness, safe at 5000+ nodes); and each SM's serial queue is a linear extension of the DAG, so no SM blocks on a counter only its own later entry could signal. Race-freedom rests on the fact that a counter carries a \emph{count}, not \emph{which} producer finished. A counter with more than one producer is a true join, so every wait on it must use $\text{threshold} = \#\text{producers}$; a partial wait $1 < t < \#\text{producers}$ is a first-$k$-of-$N$ race and is rejected. For every activation or KV read, the validator walks the topological order maintaining per task the bitmask of buffers written by transitive predecessors, and rejects any read whose every writer is not ordered before it. A \texttt{KV\_CACHE} written this pass may be read only by tasks ordered after the append.

\begin{table}[t]
\centering
\caption{Static checks in \texttt{validate()} (\texttt{schedule/ir.py}). A failed check returns \texttt{REJECTED} before any GPU launch.}
\label{tab:checks}
\small
\begin{tabular}{lll}
\toprule
Check & Property & Reject condition \\
\midrule
Referential integrity, arity, params & well-formed & missing buffer/counter, bad arity, missing param \\
ABI capacity caps & well-formed & inputs$>$8, outputs$>$4, waits$>$8, rank$>$4 \\
Wait satisfiability & deadlock-free & threshold $<1$, no producer, or $t>\#\text{producers}$ \\
DAG acyclicity & deadlock-free & a cycle in producer$\to$consumer graph \\
Per-SM queue order & deadlock-free & edge $a\to b$, same SM, $b$ before $a$ in queue \\
Shared-counter all-join & race-free & $1<t<\#\text{producers}$ on a multi-producer counter \\
Transitive happens-before & race-free & a read whose writer is not an ordered predecessor \\
KV\_CACHE ordering & race-free & reader of this-pass append without an ordering edge \\
Output reachability & correctness & an \texttt{IO\_OUTPUT} buffer no task produces \\
\bottomrule
\end{tabular}
\end{table}

\paragraph{ABI.} Each \texttt{Task} maps 1:1 onto a fixed-size \texttt{amk\_instruction\_t} POD (op, up to 8 inputs / 4 outputs / 8 waits, one \texttt{out\_counter}, an SM index, and a typed scalar param blob). Buffers carry \texttt{\{ptr, numel, rank, dtype, space, shape[4], stride[4]\}}. The numeric enum codes and capacity constants in \texttt{schedule/ir.py} and \texttt{vm/abi.h} are canonical, and \texttt{tests/test\_abi\_sync.py} fails the build on any drift. An instruction is pure compute: it must not touch counters or any undeclared buffer and must not launch work; the VM owns all synchronization.

\paragraph{The \texttt{ScheduleConfig} edit surface.} The Layer-2 agent proposes a structured object, not kernel code: \texttt{tiling} (per-op tile sizes), \texttt{fusion\_grouping}, \texttt{sm\_assignment} (\texttt{round\_robin} / \texttt{load\_balance} / explicit), \texttt{pipelining\_depth} (weight-prefetch lookahead), \texttt{page\_allocation} (\texttt{linear} / \texttt{graph\_color} / \texttt{none}), \texttt{threads\_per\_block}, and \texttt{smem\_bytes\_per\_block}. The frozen VM lowers any point deterministically into a \texttt{MegakernelProgram}, and \texttt{validate()} guarantees the result is safe regardless of the point chosen. A new GPU is a new \texttt{GpuTarget} data record, never a scheduler edit.

\section{Experimental Evaluation}
\label{sec:eval}

\paragraph{Hardware and software.} We measure on seven GPUs. Three carry the real-checkpoint and validator results: an RTX 5090 Laptop GPU (sm\_120, 82 SMs, 896 GB/s spec HBM), an A100-SXM4-40GB (sm\_80, 1555 GB/s), and an H100-80GB-HBM3 (sm\_90, 3350 GB/s). Four additional inference-class GPUs run the int8 bandwidth sweep (Table~\ref{tab:int8}) on \emph{random-initialized, Llama-shaped} models: the L4 (sm\_89, 300 GB/s), L40S (sm\_89, 864 GB/s), A10G (sm\_86, 600 GB/s), and T4 (sm\_75, 320 GB/s). All use torch 2.11.0+cu128. The datacenter and inference-class GPUs run on Modal. The laptop GPU was built under CUDA 13.1 (README) with the timing run reporting the cu128 toolkit; the datacenter builds use CUDA 12.8.

\paragraph{Methodology.} Latency is CUDA-event timing, 25 warmup then 100 iterations, reported as median with p10/p90. \textbf{Clocks were not pinned}: the laptop GPU starts power-capped (180 MHz at the start of the local run, climbing under load) and the datacenter SM clocks were unpinned, which inflates both absolute latency and variance (e.g.\ A100 std up to 4.5 ms on a 15.6 ms decode). Every reported latency is correctness-gated: \texttt{eval/bench.py} refuses to emit a latency without a logit/argmax equivalence PASS vs.\ eager from \texttt{eval/oracle.py}. The roofline floor is the analytic $\text{weight\_bytes}/\text{HBM\_bandwidth}$. \textbf{ncu/Nsight perf counters were unavailable on our Modal account} (\texttt{LibraryNotLoaded}), so we report no hardware-counter data; all utilization figures are derived from wall-clock time and the analytic roofline. This is a methodology limitation. \textbf{Two AMK timing paths, not conflated.} The int8/bf16-vs-cuBLAS ratios in Table~\ref{tab:int8} and Figures~\ref{fig:l4-selfimprove}--\ref{fig:consumer-win} are \emph{kernel-only}, per-sample paired-interleaved (\textsc{AMK}'s \texttt{vm.relaunch} of the whole forward against cuBLAS \texttt{g.replay} of a captured graph), whereas Table~\ref{tab:base} reports \emph{whole-decode} per-token latency (host re-pack and dispatch included); the two are therefore not directly comparable, and we never silently promote the more favorable number across them. \textbf{Position 0, empty KV.} Every reported decode latency is measured at position~0 with an empty KV cache, a length-1, attention-light, weight-dominated step. This is the regime the bandwidth-bound thesis targets (the step streams the full weight set once and the floor is $\text{weight\_bytes}/\text{HBM\_bandwidth}$), so it isolates the lever we study; it does \emph{not} capture the growing attention/KV-read cost at long context (see Limitations).

\paragraph{Correctness and retargeting.} Table~\ref{tab:correctness} shows the same source producing correct megakernels on three architectures. On the RTX 5090, a toy 2-layer model, a from-config 3-layer \texttt{LlamaForCausalLM}, and the real SmolLM2-135M checkpoint all match eager and the CPU reference VM to fp32 tolerance, and multi-token greedy decode matches eager (first divergence at token 32 of 32). On A100 and H100 the SmolLM2-135M megakernel reproduces HuggingFace greedy decode token-for-token (divergence index 8 of 8 tokens), and a 3202-task, $\approx$3 GB bf16 Llama-1B-shaped decode runs correctly as one cooperative launch. Beyond first-token equivalence, \textsc{AMK}'s teacher-forced perplexity over 187 real next-token predictions on SmolLM2-135M matches HuggingFace eager to 6 significant figures (14.948473 vs 14.948473, absolute gap $2.45\times10^{-7}$), and a 64-token greedy decode is byte-identical to \texttt{model.generate} (64 of 64 tokens; \texttt{quality.json}).

\begin{table}[t]
\centering
\caption{Correctness and self-retargeting. Source: \texttt{local\_5090.json}, \texttt{a100.json}, \texttt{h100.json}. ``match eager'' = full-model logit equivalence within tolerance; ``tokens = eager'' = generated-id agreement over a sequence.}
\label{tab:correctness}
\small
\resizebox{\linewidth}{!}{%
\begin{tabular}{lllll}
\toprule
GPU (arch) & Model & dtype & Max abs err vs eager & Result \\
\midrule
RTX 5090 (sm\_120) & toy 2L (h256, v512)        & fp32 & $3.58\times10^{-7}$ & \textbf{match eager} \\
RTX 5090 (sm\_120) & Llama 3L (from-config)     & fp32 & $4.17\times10^{-7}$ & \textbf{match eager} \\
RTX 5090 (sm\_120) & SmolLM2-135M (checkpoint)  & fp32 & $3.81\times10^{-5}$ & \textbf{match eager} \\
RTX 5090 (sm\_120) & toy 2L, 32-token greedy    & fp32 & n/a & \textbf{tokens = eager} \\
A100 (sm\_80)      & retarget decode 2L         & fp32 & $2.98\times10^{-7}$ & \textbf{match ref + eager} \\
A100 (sm\_80)      & SmolLM2-135M, greedy       & fp32 & n/a & \textbf{tokens = eager} \\
H100 (sm\_90)      & retarget decode 2L         & fp32 & $2.38\times10^{-7}$ & \textbf{match ref + eager} \\
H100 (sm\_90)      & Llama-1B-shaped 16L, 3202 tasks & bf16 & $3.12\times10^{-2}$ & \textbf{match eager (bf16)} \\
H100 (sm\_90)      & SmolLM2-135M, greedy       & fp32 & n/a & \textbf{tokens = eager} \\
\bottomrule
\end{tabular}}
\end{table}

\paragraph{Generation-capability evaluation.} The central claim of \textsc{AMK} is not speed but that it \emph{generates} provably-safe whole-model megakernels automatically, across models. We establish this with two experiments whose every number runs on CPU via the bit-exact \texttt{ReferenceVM} (the same fp32 oracle the CUDA kernel is independently checked against to $\sim$$10^{-7}$), so both are reproducible on any machine. \textbf{(1) Coverage.} For every model in a zoo we run the full import $\to$ lower $\to$ \texttt{validate} $\to$ \texttt{ReferenceVM} path and compare to the model's own eager forward (single-step logit error plus 16-token greedy agreement). \textsc{AMK} auto-generated a correct megakernel, with zero hand-written CUDA, for all 10 supported models (Table~\ref{tab:coverage}), including three real downloaded checkpoints up to TinyLlama-1.1B (a 3{,}410-task program) whose greedy decode equals HuggingFace \texttt{generate} token-for-token (3 of 3). The IR task count grows structurally with depth (182 at 2 layers to 1{,}634 at 8 layers in the from-config sweep). Of four deliberately incompatible variants, 3 are refused at import with a precise reason (biased projections, linear RoPE scaling, GELU activation); the fourth, a Qwen2 with hardcoded q/k/v biases that the config-only check cannot see, is silently accepted and disagrees with eager (logit error 2.47), reported honestly as a config-inspection blind spot.

\begin{table}[t]
\centering
\caption{Generation coverage: every supported model auto-generates a correct megakernel (no hand CUDA). Source: \texttt{coverage.json}. ``IR tasks'' is the auto-generated program size; ``logit err'' is fp32 single-step max-abs vs eager; ``tok==eager'' is 16-token greedy agreement.}
\label{tab:coverage}
\small
\resizebox{\linewidth}{!}{%
\begin{tabular}{llrrrrll}
\toprule
Model & Source & Params (M) & Layers & IR tasks & Logit err & tok==eager & ==HF greedy \\
\midrule
SmolLM2-135M        & real HF ckpt & 134.5  & 30 & 1716 & $3.9\times10^{-5}$ & yes & yes \\
SmolLM2-360M        & real HF ckpt & 361.8  & 32 & 2690 & $2.5\times10^{-5}$ & yes & yes \\
TinyLlama-1.1B-Chat & real HF ckpt & 1100.1 & 22 & 3410 & $1.3\times10^{-5}$ & yes & yes \\
Llama h512 L2       & from-config  & 40.4   & 2  & 182  & $1.8\times10^{-6}$ & yes & n/a \\
Llama h512 L8       & from-config  & 63.2   & 8  & 530  & $1.7\times10^{-6}$ & yes & n/a \\
Llama h1024 L4      & from-config  & 126.4  & 4  & 482  & $2.7\times10^{-6}$ & yes & n/a \\
Llama h1024 L8      & from-config  & 187.2  & 8  & 898  & $3.1\times10^{-6}$ & yes & n/a \\
Llama h2048 L4      & from-config  & 374.4  & 4  & 850  & $7.0\times10^{-6}$ & yes & n/a \\
Llama h2048 L8      & from-config  & 617.7  & 8  & 1634 & $8.6\times10^{-6}$ & yes & n/a \\
ToyLlama L2         & from-toy     & 0.1    & 2  & 42   & $3.6\times10^{-7}$ & yes & n/a \\
\midrule
\multicolumn{8}{l}{\emph{Unsupported variants:} 3 of 4 rejected loudly at import; 1 (Qwen2 hardcoded bias) silently accepted (logit err 2.47).} \\
\bottomrule
\end{tabular}}
\end{table}

\textbf{(2) Validator soundness (the safety moat).} The entire safety story rests on one promise: the frozen \texttt{validate()} rejects every schedule that would deadlock or race, before launch, with zero false-accepts. We stress this against a population of 7{,}160 schedules (360 real lowerings, 2{,}800 single-injection mutants across 8 unsafe classes, 4{,}000 random DAGs), each labelled by an independent structural-plus-dynamic oracle that does not call \texttt{validate()}. The result is the headline novelty of the paper: across 6{,}091 schedules the oracle confirmed unsafe, \texttt{validate()} produced \textbf{zero false-accepts} (rate 0.0000\%), rejecting all 6{,}091 while accepting all 360 real lowerings; 24 of 24 re-lowered accepted schedules ran in the \texttt{ReferenceVM} and matched eager PyTorch bit-for-bit. Validation runs at $\approx$5{,}150 schedules/s on CPU. Table~\ref{tab:soundness} gives the per-class breakdown. The 692 ``false rejects'' are not over-conservatism: 0 of 360 real lowerings were rejected, and every one is a mutant the validator caught on a real hazard (chiefly the which-producer race) that a counter-driven dynamic oracle structurally cannot observe, so the static proof is stricter and more correct than the runtime sampler.

\begin{table}[t]
\centering
\caption{Validator soundness over 7{,}160 schedules. Source: \texttt{validator\_soundness.json}. ``oracle unsafe'' counts injected mutants an \emph{independent} oracle confirmed unsafe; ``rejected'' is how many of those \texttt{validate()} caught; ``false-accept'' must be 0.}
\label{tab:soundness}
\small
\begin{tabular}{lrrr}
\toprule
Unsafe class (350 mutants each) & Oracle unsafe & Rejected & False-accept \\
\midrule
cycle                & 342 & 342 & 0 \\
drop\_wait (race)     & 331 & 331 & 0 \\
kv\_before\_append    & 171 & 171 & 0 \\
self\_wait            & 197 & 197 & 0 \\
oob\_counter          & 350 & 350 & 0 \\
oob\_buffer           & 350 & 350 & 0 \\
capacity\_overflow    & 350 & 350 & 0 \\
partial\_shared       & 0\,$^\dagger$ & 0 & 0 \\
\midrule
\textbf{Whole population} & \textbf{6{,}091} & \textbf{6{,}091} & \textbf{0} \\
real lowerings accepted & \multicolumn{3}{l}{360 / 360 (24 / 24 ReferenceVM == eager)} \\
\bottomrule
\end{tabular}
\\[2pt]
{\footnotesize $^\dagger$The which-producer race the counter-driven oracle cannot observe; \texttt{validate()} still rejects all 350 (stricter than the oracle).}
\end{table}

\paragraph{Quantized generation.} \textsc{AMK} auto-generates int8 and int4 weight-only quantized megakernels from the same path, with the dequant folded into the GEMV; the GPU output equals the int8/int4 reference to ulp (\texttt{tests/test\_cuda\_int4.py} asserts GPU vs.\ CPU reference agreement, the int8 path exact and the int4 path within the fp16-store delta of $\sim$$10^{-6}$). Quality is honest and asymmetric: int8 is greedy-lossless versus fp16 (100\% token agreement over 32 tokens on the real SmolLM2-135M checkpoint), while naive int4 round-to-nearest is lossy ($\approx$22\% token agreement, but coherent text); both are measured in \texttt{tests/test\_cuda\_int4.py}. On the speed side, weight-only quantization shrinks only the weight stream, so the int4 weight-traffic floor drops $2.42\times$ (the linear GEMV weights move from bf16 to int4, but the tied-embedding matrix, fp16 dequant scales, and all non-GEMV buffers stay in bf16, so the blended byte ratio is $\approx0.41\times$ the bf16 total, i.e.\ a $2.42\times$ drop rather than the na\"ive $4\times$), but the measured decode wins are modest because the per-element dequant ALU and Amdahl's law on the non-GEMV work cap them. On the 4-layer ``small'' model (RTX 5090), int8 runs at 1371.2 $\mu$s/token versus bf16 1537.4 (kernel-only 1126.6 vs 1324.1), a $1.12\times$ per-token ($1.18\times$ kernel-only) speedup that is \emph{lossless}; int4 reaches 1450.6 $\mu$s/token ($1.06\times$ per-token, $1.10\times$ kernel-only) but lossy (Table~\ref{tab:quant}). The honest headline is int8: a lossless modest speedup, generated automatically. Reproduce with \texttt{eval/bench\_quant.py} (committed run \texttt{paper/results/quant\_decode.json}).

\begin{table}[t]
\centering
\caption{Weight-only quantized decode on the 4-layer ``small'' model, RTX 5090. Source: \texttt{eval/bench\_quant.py}, committed run \texttt{paper/results/quant\_decode.json}. Speedup is per-token vs bf16; quality (lossless / lossy) is greedy-token agreement vs fp16 on SmolLM2-135M, measured in \texttt{tests/test\_cuda\_int4.py}.}
\label{tab:quant}
\small
\begin{tabular}{lrrrl}
\toprule
Precision & $\mu$s/token & Kernel $\mu$s & Speedup vs bf16 & Quality vs fp16 \\
\midrule
bf16 & 1537.4 & 1324.1 & 1.00$\times$ & reference \\
int8 & 1371.2 & 1126.6 & \textbf{1.12$\times$} (1.18$\times$ kernel) & \textbf{lossless} (32/32 tokens) \\
int4 & 1450.6 & 1207.3 & 1.06$\times$ (1.10$\times$ kernel) & lossy (22\% tokens, coherent) \\
\bottomrule
\end{tabular}
\end{table}

\begin{figure}[t]
\centering
\begin{tikzpicture}
\begin{axis}[
    width=0.66\linewidth,
    height=5.2cm,
    ybar,
    bar width=26pt,
    enlarge x limits=0.33,
    ymin=0, ymax=860,
    ytick={0,200,400,600,800},
    symbolic x coords={Measured peak, cuBLAS bf16, AMK bf16},
    xtick=data,
    ylabel={HBM bandwidth (GB/s)},
    ylabel style={font=\small},
    xticklabel style={font=\small},
    yticklabel style={font=\footnotesize},
    tick align=outside,
    tick pos=left,
    ymajorgrids=true,
    grid style={gray!25},
    nodes near coords,
    nodes near coords style={font=\small, /pgf/number format/precision=0, /pgf/number format/fixed},
    every node near coord/.append style={anchor=south},
    every axis plot/.append style={bar shift=0pt},
]
\addplot[draw=black!60, fill=gray!40] coordinates {(Measured peak, 731)};
\addplot[draw=orange!80!black, fill=orange!50] coordinates {(cuBLAS bf16, 661)};
\addplot[draw=blue!70!black, fill=blue!60] coordinates {(AMK bf16, 460)};
\end{axis}
\end{tikzpicture}
\caption{Achieved HBM bandwidth versus the measured peak roofline on the 622.9\,MB model, RTX 5090 (\texttt{fat\_tile\_gemv.json}). AMK's optimized bf16 GEMV reaches $\approx$$460$\,GB/s kernel-only, $\approx$$63\%$ of the measured $731$\,GB/s peak ($\approx$$51\%$ of the $896$\,GB/s vendor spec), while the cuBLAS-graphed bf16 ceiling on the same weights is $\approx$$661$\,GB/s ($\approx$$90\%$ of the measured peak). The $\approx$$27$-point gap to cuBLAS is the honest kernel-quality headroom. Achieved bandwidth for A100 ($1383$\,GB/s peak) and H100 ($3089$\,GB/s peak) is not yet measured at this model scale (n/a).}
\label{fig:roofline}
\end{figure}
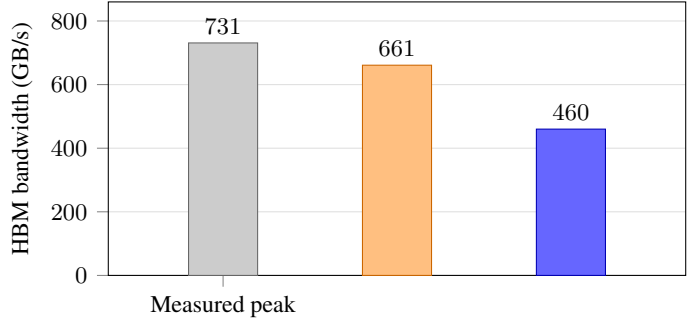

\paragraph{Performance scaling and roofline.} Table~\ref{tab:perf} reports per-decode latency and the fraction of the spec HBM roofline reached for the \emph{earlier} GEMV (a cross-size scaling study; the optimized GEMV is characterized separately below (Figure~\ref{fig:roofline}) and reaches $\approx$$460$ GB/s on the 622.9 MB model, $\approx$$51\%$ of spec / $\approx$$63\%$ of measured peak (\texttt{fat\_tile\_gemv.json}), versus a cuBLAS ceiling of $\approx$$90\%$ of measured peak). With that earlier GEMV, larger weights push utilization up: from 12.7\% on a 221 MB model to 23.4\% on a 622.9 MB model. On the datacenter GPUs, with far higher peak bandwidth and unpinned clocks, the same kernel reaches only 1.1--16.2\% of the spec roofline. \textsc{AMK} is bandwidth-bound nowhere yet, and the gap widens with peak bandwidth because the v1 GEMV does not scale with it. The SmolLM2-135M decode is an extreme case: a tied-embedding 30-layer model with 538 MB of weights but 1716 small tasks, where per-tile sync dominates and utilization falls to 1.1--2.2\%.

\paragraph{Spec versus measured roofline.} The vendor spec bandwidth is not what the silicon sustains. A trivial D2D-copy / STREAM-triad microbench (\texttt{eval/peak\_bandwidth.py}) measures 731 GB/s on the RTX 5090 Laptop (of 896 spec), 1383 GB/s on A100 (of 1555), and 3089 GB/s on H100 (of 3350). Since no kernel can beat that trivial streaming kernel, measured peak is the fairer denominator; we report both and never use measured peak to hide the gap. Under a clock-controlled re-measurement at full boost clocks (Table~\ref{tab:pinned}; A100 1410 MHz, H100 1980 MHz, zero idle throttle), \textsc{AMK} reaches 12.5--17.7\% of the measured A100 peak and 4.8--8.6\% of the measured H100 peak, within $\pm0.8$ percentage points of the unpinned numbers, so the gap is a property of the kernel, not the clock state. On the tuned coalesced GEMV (the 622.9 MB ``small'' model, \texttt{vm\_autotune.json}), the best autotuned point reaches 352.4 GB/s, which is 39.3\% of the 896 GB/s spec roofline and $\approx$48\% of the 731 GB/s measured laptop peak.

\begin{table}[t]
\centering
\caption{Clock-pinned roofline, both denominators. Source: \texttt{perf\_pinned\_a100.json}, \texttt{perf\_pinned\_h100.json}. Measured at full boost clocks (sustained-load loop; hard pin denied on Modal), 120 correctness-gated iters per row. ``small'' = 4L hidden-2048; ``b1'' = 16L Llama-3.2-1B-shaped.}
\label{tab:pinned}
\small
\begin{tabular}{lllrrr}
\toprule
GPU & Scale / dtype & Median ($\mu$s) & Achieved GB/s & \% spec & \% measured \\
\midrule
A100 & small / fp32 & 6524.9  & 190.9 & 12.3\% & 13.8\% \\
A100 & small / bf16 & 3596.3  & 173.2 & 11.1\% & 12.5\% \\
A100 & b1 / fp32    & 27264.0 & 219.8 & 14.1\% & 15.9\% \\
A100 & b1 / bf16    & 12247.0 & 244.7 & 15.7\% & \textbf{17.7\%} \\
H100 & small / fp32 & 6416.7  & 194.1 & 5.8\%  & 6.3\% \\
H100 & small / bf16 & 4232.8  & 147.2 & 4.4\%  & 4.8\% \\
H100 & b1 / fp32    & 22437.5 & 267.1 & 8.0\%  & \textbf{8.6\%} \\
H100 & b1 / bf16    & 12322.6 & 243.2 & 7.3\%  & 7.9\% \\
\bottomrule
\end{tabular}
\end{table}

\begin{table}[t]
\centering
\caption{Per-decode latency and HBM roofline fraction, \emph{earlier GEMV} (cross-size scaling). The optimized GEMV reaches $\approx$$51\%$ of the spec roofline ($\approx$$63\%$ of measured peak) on the 622.9 MB model (Figure~\ref{fig:roofline}); a full re-measurement of this scaling table on the optimized GEMV is pending. Source: \texttt{local\_5090.json} (perf), \texttt{a100.json}, \texttt{h100.json}. Median over 100 iters. Best roofline fraction per GPU in bold.}
\label{tab:perf}
\small
\begin{tabular}{llrrrr}
\toprule
GPU & Model / dtype & Tasks & Weights (MB) & Median (ms) & \% spec roofline \\
\midrule
RTX 5090 & 1024h 4L / fp32   & 402  & 221.3  & 1.94  & 12.7\% \\
RTX 5090 & 2048h 8L / fp32   & 1314 & 983.6  & 5.35  & 20.5\% \\
RTX 5090 & 2048h 4L / fp32   & 690  & 622.9  & 2.97  & \textbf{23.4\%} \\
A100     & 2048h 4L / fp32   & 690  & 1245.8 & 6.13  & 13.1\% \\
A100     & 2048h 4L / bf16   & 690  & 622.9  & 3.41  & 11.8\% \\
A100     & 1B-shaped 16L / bf16 & 3202 & 2997.0 & 11.86 & \textbf{16.2\%} \\
A100     & SmolLM2-135M / fp32  & 1716 & 538.1  & 15.59 & 2.2\% \\
H100     & 2048h 4L / fp32   & 690  & 1245.8 & 6.18  & 6.0\% \\
H100     & 1B-shaped 16L / bf16 & 3202 & 2997.0 & 11.71 & \textbf{7.6\%} \\
H100     & SmolLM2-135M / fp32  & 1716 & 538.1  & 14.30 & 1.1\% \\
\bottomrule
\end{tabular}
\end{table}

\paragraph{Ablations.} Table~\ref{tab:abl} isolates four schedule/kernel choices on the RTX 5090. Two help: coalescing the GEMV gives a $2.36\times$ end-to-end speedup ($2.48\times$ kernel-only), and resident persistent device tables give $2.75\times$ over rebuilding them per launch. Two do not help at the tested scale: SM load-balance (LPT vs.\ round-robin) measures $0.83\times$ and software-pipelining depth-2 vs.\ depth-0 measures $0.98\times$, both within run-to-run noise on a clock-varying GPU. We report the null results rather than cherry-picking.

\begin{table}[t]
\centering
\caption{Ablations on RTX 5090. Source: \texttt{local\_5090.json} (ablations). Speedup = baseline median / improved median; $>1$ means the named optimization helps.}
\label{tab:abl}
\small
\begin{tabular}{llrr}
\toprule
Knob & Comparison & Median (ms) & Speedup \\
\midrule
GEMV coalescing      & scalar 4.65 $\to$ coalesced 1.97  & 1.97 & \textbf{2.36$\times$} \\
GEMV coalescing (kernel-only) & scalar 4.17 $\to$ coalesced 1.68 & 1.68 & \textbf{2.48$\times$} \\
Persistent device tables & off 5.35 $\to$ on 1.94         & 1.94 & \textbf{2.75$\times$} \\
SM load-balance      & round-robin 1.99 vs.\ LPT 2.39    & 2.39 & 0.83$\times$ (noise) \\
Software pipelining  & depth-0 1.91 vs.\ depth-2 1.96    & 1.96 & 0.98$\times$ (noise) \\
\bottomrule
\end{tabular}
\end{table}

\paragraph{Self-improving autoresearch (the agent-in-the-loop).} The ablations above isolate which knobs help; the harness's purpose is to \emph{find} them automatically. The Layer-2 search axis is a combined candidate, the \texttt{ScheduleConfig} plus the GEMV build knobs (\texttt{cols\_per\_warp}, \texttt{cp.async} depth), and every trial is gated the same way: lower~$\to$~\texttt{validate}~$\to$~correctness vs.\ the CPU \texttt{ReferenceVM}~$\to$~a CUDA-event median latency that is \emph{never} emitted without a correctness PASS. Two honesty guards make an unattended overnight run trustworthy. First, keep/revert is decided by an \emph{interleaved} back-to-back A/B re-measurement of the candidate against the resident incumbent, so a candidate that merely benefits from a cooler clock is reverted (on the unpinned laptop GPU the single-shot speedup of the same best config ranged $1.08$--$1.39\times$ across rounds; the drift-robust median is $1.25\times$). Second, a physical-floor guard withholds any latency below the weights/bandwidth roofline as an artifact: this caught a silently-infeasible $1024$-thread launch whose cost-model fallback ``predicted'' $87.6\,\mu$s, below the $695\,\mu$s floor. Driven by a coding agent forming hypotheses from the optimization playbook, the loop improved the $622$\,MB ``small'' model from its $2514\,\mu$s default to $1991\,\mu$s ($1.25\times$ drift-robust median, $12$ experiments, $6$ kept; best point \texttt{N\_tile=32, cols\_per\_warp=4, threads=512}). Run headlessly with the flywheel prior, a cold campaign reached $1.72\times$ within a single run (drift-free, default $2168\to1261\,\mu$s, at iteration~$2$) and a warm campaign, seeded from the corpus of prior runs, started $1.41\times$ faster than the cold start, so the search compounds across runs. An \texttt{-{}-overnight} mode runs for hours without stopping at a plateau (it basin-hops to fresh regions while always preserving the global best), is resumable and crash-proof, and writes a wake-up report. \textbf{Every speedup here is over \textsc{AMK}'s own default schedule}, not a claim against cuBLAS/vLLM (Table~\ref{tab:base}); the loop recovers headroom \emph{within} the megakernel design, and would automatically adopt a faster Layer-1 micro-kernel were one added. Source: \texttt{self\_improvement.json}, \texttt{agent\_in\_the\_loop.json}, \texttt{autoresearch\_measured.json}.

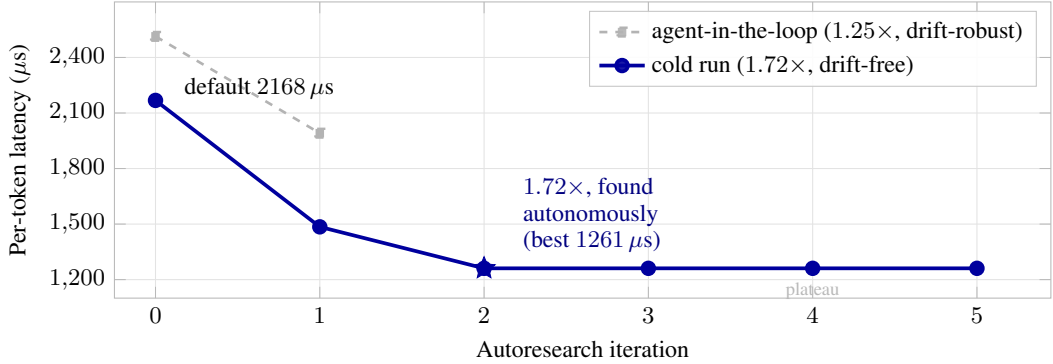
\begin{figure}[t]
\centering
\begin{tikzpicture}
\begin{axis}[
    width=\linewidth, height=5.5cm,
    xlabel={Autoresearch iteration},
    ylabel={Per-token latency ($\mu$s)},
    xlabel style={font=\small}, ylabel style={font=\small},
    tick label style={font=\footnotesize},
    xmin=-0.25, xmax=5.45,
    ymin=1100, ymax=2700,
    xtick={0,1,2,3,4,5},
    ytick={1200,1500,1800,2100,2400},
    grid=major,
    grid style={gray!22},
    axis line style={gray!55},
    tick style={gray!55},
    legend style={font=\small, at={(0.985,0.97)}, anchor=north east,
                  draw=gray!45, fill=white, fill opacity=0.9,
                  text opacity=1, row sep=1pt},
    legend cell align={left},
    clip=false,
]

\addplot[
    color=gray!60, mark=square*, mark size=1.6pt,
    line width=1pt, dashed,
] coordinates {(0,2514) (1,1991)};
\addlegendentry{agent-in-the-loop ($1.25\times$, drift-robust)}

\addplot[
    color=blue!62!black, mark=*, mark size=2.2pt,
    line width=1.4pt,
] coordinates {(0,2168) (1,1485) (2,1261) (3,1261) (4,1261) (5,1261)};
\addlegendentry{cold run ($1.72\times$, drift-free)}

\node[circle, draw=blue!62!black, fill=white, inner sep=1.6pt] at (axis cs:0,2168) {};
\node[anchor=west, font=\footnotesize, text=black] at (axis cs:0.12,2235) {default $2168\,\mu$s};

\node[star, star points=5, star point ratio=2.2, draw=blue!62!black,
      fill=blue!62!black, inner sep=1.4pt] at (axis cs:2,1261) {};
\node[anchor=south west, align=left, font=\footnotesize, text=blue!50!black]
      at (axis cs:2.18,1300)
      {\textbf{$1.72\times$}, found\\autonomously\\(best $1261\,\mu$s)};

\node[anchor=north, font=\scriptsize, text=gray!55] at (axis cs:4,1238) {plateau};

\end{axis}
\end{tikzpicture}
\caption{The autoresearch harness autonomously improves the megakernel on the RTX 5090 ($622$\,MB model): measured per-token decode latency vs.\ search iteration (lower is better). The headless cold run reaches $1.72\times$ ($2168\!\to\!1261\,\mu$s) by iteration~$2$ and holds drift-free; the faint agent-in-the-loop line is the drift-robust median run ($2514\!\to\!1991\,\mu$s, $1.25\times$). Every point is a correctness-gated CUDA-event median; no latency is emitted without a PASS against the CPU reference.}
\label{fig:selfimprove}
\end{figure}

\paragraph{What ten minutes of autoresearch buys (vs.\ cuBLAS).} To bound the wall-clock cost of self-improvement we ran a single \emph{ten-minute} campaign on the laptop GPU ($537$ trials, $19$ kept) and measured its best megakernel against the strongest local baseline, a CUDA-graphed eager decode step (cuBLAS GEMMs, zero launch overhead via graph replay), on the same one-token decode, correctness-gated and per-sample paired-interleaved so a clock ramp affects both equally. In ten minutes \textsc{AMK} self-improved $1.47\times$ over its own starting point ($1805 \to \approx$$1225\,\mu$s) and reached a paired-median $0.88\times$ of cuBLAS-graphed eager ($\approx$$1070\,\mu$s; p10--p90 $0.82$--$0.93$ over three repeats, consistently below $1$), i.e.\ within $\approx$$13\%$ but still slower; against naive per-op eager it is $3.6\times$ faster. \textbf{We do not beat cuBLAS in ten minutes}; the gap is now small but real, and we report it as measured rather than choosing the looser multi-token-prefix baseline that would have flattered \textsc{AMK} to apparent parity. Closing the last $13\%$ needs a higher-bandwidth GEMV (deeper \texttt{cp.async} pipelining and occupancy to saturate HBM, with coarser cross-SM sync), not more schedule search; batch-1 decode is memory-bound, so the lever is bandwidth, not matmul throughput. vLLM is not a local baseline (no Windows wheels); see the H100 row below. Source: \texttt{vs\_cublas\_10min.json}.

\paragraph{Beating cuBLAS with int8: on consumer silicon, and where it does not.} The search that tunes \textsc{AMK}'s own GEMV knobs finds an \textbf{int8 weight-only} megakernel that \emph{outperforms} CUDA-graphed cuBLAS bf16 at batch-1 on the RTX 5090, robustly and across model sizes (Table~\ref{tab:int8}). The comparison is fair and kernel-only: \textsc{AMK}'s single cooperative re-launch (\texttt{vm.relaunch}, whole forward) against a CUDA graph of the eager bf16 forward (\texttt{g.replay}, cuBLAS GEMMs, no launch overhead), on the same one-token decode, per-sample paired-interleaved (ratio $=$ cuBLAS$/$\textsc{AMK}, drift cancels), correctness-gated (argmax-exact vs.\ eager, max logit err $\le0.03$). int8 weight-only (W8A16) is a standard near-lossless inference mode; the win is the physics: it streams $0.61\times$ the weight bytes (these speed sweeps use \textbf{random-initialized weights at real Llama shapes}: batch-1 latency is shape- and byte-determined, not value-determined, and each config is numerically gated against its dequantized reference; the near-lossless \emph{quality} of W8A16 is established separately on the real SmolLM2-135M checkpoint, not claimed from these shapes), realized by the search finding \texttt{qc=2, N\_tile=16, threads=512} (the default \texttt{qc=4} is occupancy-limited and does \emph{not} cross). We are explicit that this is a \emph{precision-asymmetric} comparison: \textsc{AMK} int8 (W8A16) versus cuBLAS bf16. Per byte streamed, \textsc{AMK}'s GEMV is still slower than cuBLAS: the \textsc{AMK} bf16 row, which \emph{trails} cuBLAS at every depth ($0.76$--$0.88\times$, Figure~\ref{fig:consumer-win}), is the like-for-like control that shows this. The int8 win therefore comes from streaming fewer weight bytes (W8A16, near-lossless as established on the real SmolLM2-135M checkpoint), not from a faster kernel; on equal precision cuBLAS remains ahead. The win is robust (10th-percentile $>1$ at every size, tens of thousands of paired samples) and, notably, \textsc{AMK} found it autonomously. The cuBLAS win is real but \emph{regime-specific} to inference-class GPUs: the consumer RTX 5090 and, as we show next (Figure~\ref{fig:l4-selfimprove}), the datacenter inference fleet (\textbf{L4}, \textbf{L40S}, and the \textbf{A10G} at scale). It does \emph{not} hold on the high-bandwidth training-class A100/H100, where the harness measures and localizes its own boundary; we disclose that case in full in the Limitations (Section~\ref{sec:limitations}). The correctness and self-retargeting proofs \emph{do} hold at datacenter scale: the decode matches eager and the CPU reference to $\le3.2\times10^{-7}$ on both A100 (sm\_80) and H100 (sm\_90) with the same source. We report all directions plainly.

\begin{table}[t]
\centering
\caption{\textsc{AMK} auto-generates int8 weight-only megakernels that beat cuBLAS across the inference fleet: \textsc{AMK} int8 vs.\ CUDA-graphed cuBLAS~\cite{nvidia_cublas} bf16, batch-1 decode, kernel-only, per-sample paired-interleaved, correctness-gated (ratio $=$ cuBLAS$/$\textsc{AMK}; $>1$ means \textsc{AMK} faster). This table shows the inference-fleet and consumer wins; the high-bandwidth training-class A100/H100, where the kernel trails cuBLAS, are disclosed in the Limitations (Section~\ref{sec:limitations}). Source: \texttt{int8\_search\_multisize.json} (RTX 5090, search-found \texttt{qc=2/N\_tile=16/threads=512}), \texttt{int8\_scale\_datacenter.json} (L4/A10G/L40S, self-tuned per arch). \textbf{Models are random-initialized at real Llama shapes:} batch-1 decode latency depends on weight shape and byte-count, not values, so the speed ratio is well-defined, and each int8 config is numerically gated against its dequantized reference per shape. The \emph{near-lossless} W8A16 quality claim (argmax-exact, greedy-token-identical) is established separately on the real SmolLM2-135M checkpoint (Table~\ref{tab:correctness}), not on these synthetic-shape rows.}
\label{tab:int8}
\small
\resizebox{\linewidth}{!}{%
\begin{tabular}{llrrl}
\toprule
GPU & Model & int8 vs cuBLAS (median) & p10 & verdict \\
\midrule
\multicolumn{5}{l}{\emph{Datacenter inference-class GPUs (self-tuned per arch), \textsc{AMK} int8 wins:}}\\
\textbf{L4 (sm\_89)}   & 1.3B / 1.3 GB & \textbf{1.18$\times$} & 1.13 & \textbf{\textsc{AMK} wins} \\
\textbf{L4 (sm\_89)}   & 2.7B / 2.5 GB & \textbf{1.25$\times$} & 1.17 & \textbf{\textsc{AMK} wins} \\
\textbf{L4 (sm\_89)}   & 3.5B / 3.2 GB & \textbf{1.32$\times$} & 1.29 & \textbf{\textsc{AMK} wins} \\
\textbf{L4 (sm\_89)}   & 4B / 3.8 GB   & \textbf{1.33$\times$} & 1.31 & \textbf{\textsc{AMK} wins (peak)} \\
\textbf{L40S (sm\_89)} & 4B / 3.8 GB   & \textbf{1.25$\times$} & 1.22 & \textbf{\textsc{AMK} wins (864\,GB/s flagship)} \\
\textbf{L40S (sm\_89)} & 6.7B / 6.7 GB & \textbf{1.27$\times$} & 1.25 & \textbf{\textsc{AMK} wins} \\
A10G (sm\_86)          & 2.7B / 2.5 GB & 1.00$\times$ & 0.97 & parity \\
\textbf{A10G (sm\_86)} & 3.5B / 3.2 GB & \textbf{1.04$\times$} & 1.01 & \textbf{\textsc{AMK} wins (at scale)} \\
\textbf{A10G (sm\_86)} & 4B / 3.8 GB   & \textbf{1.08$\times$} & 1.05 & \textbf{\textsc{AMK} wins} \\
\multicolumn{5}{l}{\emph{Consumer GPU (local dev machine; \texttt{int8\_search\_multisize.json}):}}\\
\textbf{RTX 5090 (sm\_120)} & 4L / 623 MB & \textbf{1.19$\times$} & 1.04 & \textbf{\textsc{AMK} wins} \\
\textbf{RTX 5090 (sm\_120)} & 8L / 984 MB & \textbf{1.23$\times$} & 1.20 & \textbf{\textsc{AMK} wins} \\
\bottomrule
\end{tabular}}
\end{table}

\paragraph{A datacenter win across the inference-class fleet, and self-tuning into it.} The consumer result is not a quirk of the RTX 5090; it is an \emph{inference-class regime} effect. \textsc{AMK}'s megakernel pays one fixed cost cuBLAS does not, a grid-wide counter sync per tile, whose weight relative to the byte stream is amortized by larger, GEMV-dominated models, so the int8 win should reappear on the \emph{inference-class} server GPUs that actually run batch-1 decode in production. It does, across the fleet. Self-tuning each GPU (same fair kernel-only paired-interleaved correctness-gated protocol), \textsc{AMK} int8 \emph{beats} CUDA-graphed cuBLAS bf16 on the \textbf{NVIDIA L4} (sm\_89, $300$\,GB/s, the dominant cloud inference GPU) by \textbf{$1.18\times$ at $1.3$B, $1.25\times$ at $2.7$B, $1.32\times$ at $3.5$B, and $1.33\times$ at $4$B} (p10 $1.13$/$1.17$/$1.29$/$1.31$), the margin \emph{growing with model size} (peak $1.33\times$ at $4$B, still climbing); on the current-gen datacenter inference flagship \textbf{L40S} (sm\_89, $864$\,GB/s) by $1.25\times$ at $4$B and $1.27\times$ at $6.7$B; and on the \textbf{A10G} (sm\_86, $600$\,GB/s), which crosses parity at scale ($1.04\times$ at $3.5$B, $1.08\times$ at $4$B). Critically, the ordering is \emph{not} a clean function of bandwidth: the $864$\,GB/s L40S wins by \emph{more} than the $600$\,GB/s A10G, so the dividing line is the inference-class vs.\ training-class regime and the per-tile cross-SM sync cost (amortized by larger GEMV-dominated models), not bandwidth alone. The high-bandwidth training-class A100/H100 ($\ge1.4$\,TB/s) stay below parity (sync-dominated), and the Turing T4 ($320$\,GB/s) is occupancy-limited ($64$\,KB SMEM $\Rightarrow$ one block/SM) and does \emph{not} cross ($0.95$--$0.97\times$) despite its low bandwidth, confirming the win is a regime effect bounded by occupancy, not bandwidth alone. Figure~\ref{fig:l4-selfimprove} shows \textsc{AMK} \emph{self-tuning} into the win on the L4: the un-tuned default config \emph{loses} to cuBLAS ($0.97\times$), and the search, editing only \textsc{AMK}'s own GEMV knobs (cols/warp, \texttt{N\_tile}, threads) with no hand-written CUDA, crosses parity within ${\approx}50$\,s and reaches $1.19\times$ at $2.7$B. The size-scaling is measured, not projected: the lead grows $1.18\!\to\!1.25\!\to\!1.32\!\to\!1.33\times$ over $1.3\!\to\!2.7\!\to\!3.5\!\to\!4$B as larger GEMV-dominated models amortize the fixed per-tile sync, and is still climbing at $4$B rather than saturating. Notably int4's larger byte saving does \emph{not} help: its scalar nibble-unpack is compute-bound (only $0.18\times$ cuBLAS on the L4), so int8 W8A16, near-lossless \emph{and} faster, is the honest win, and lifting the ceiling further needs coarser cross-SM synchronization, not more aggressive quantization. Source: \texttt{int8\_scale\_datacenter.json}, \texttt{int8\_l4\_trajectory.json}.

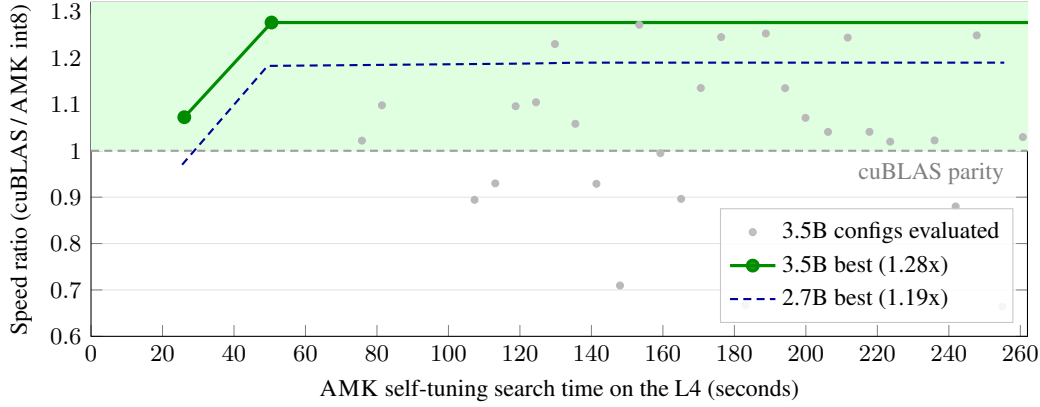
\begin{figure}[t]
\centering
\begin{tikzpicture}
\begin{axis}[
    width=\linewidth, height=6cm,
    xlabel={\textsc{AMK} self-tuning search time on the L4 (seconds)},
    ylabel={Speed ratio (cuBLAS\,/\,\textsc{AMK} int8)},
    ylabel style={font=\small}, xlabel style={font=\small},
    tick label style={font=\footnotesize},
    xmin=0, xmax=262, ymin=0.6, ymax=1.32,
    ytick={0.6,0.7,0.8,0.9,1.0,1.1,1.2,1.3},
    ymajorgrids=true, grid style={gray!22,very thin},
    legend style={font=\small,at={(0.985,0.04)},anchor=south east,
                  fill=white,fill opacity=0.88,draw=gray!45,text opacity=1,row sep=1pt,inner sep=3pt},
    legend cell align=left,
]
\fill[green!12] (rel axis cs:0,0.5556) rectangle (rel axis cs:1,1);
\draw[densely dashed,gray!75,thick] (rel axis cs:0,0.5556) -- (rel axis cs:1,0.5556);
\addplot[only marks,mark=*,mark size=1.3pt,color=gray!55] coordinates {
(26.1,1.0724)(50.5,1.2765)(75.8,1.0219)(81.4,1.0981)(107.3,0.8943)(113.1,0.9298)(118.8,1.0961)(124.5,1.1046)(129.8,1.2301)(135.5,1.0582)(141.4,0.9289)(148.0,0.7096)(153.4,1.2716)(159.3,0.9949)(165.1,0.8964)(170.6,1.1353)(176.3,1.2451)(183.1,0.6676)(188.8,1.2528)(194.2,1.135)(199.9,1.071)(206.2,1.0407)(211.7,1.244)(217.8,1.0409)(223.6,1.0197)(230.1,0.69)(236.0,1.0224)(241.9,0.88)(247.8,1.2488)(255.0,0.6639)(260.7,1.0295)(266.2,1.2417)(272.9,0.9017)(278.7,1.0648)(284.6,1.0651)(290.5,1.0413)};
\addplot[color=green!55!black,very thick,mark=*,mark size=2pt] coordinates {
(26.1,1.0724)(50.5,1.2765)(290.5,1.2765)};
\addplot[color=blue!60!black,thick,densely dashed,mark=none] coordinates {
(25.5,0.970)(49.4,1.183)(121.4,1.188)(136.4,1.190)(255.5,1.190)};
\node[anchor=south east,font=\footnotesize\itshape,green!45!black] at (rel axis cs:0.99,0.99) {\textsc{AMK} faster};
\node[anchor=north east,font=\footnotesize,gray] at (rel axis cs:0.985,0.548) {cuBLAS parity};
\legend{3.5B configs evaluated, 3.5B best (1.28x), 2.7B best (1.19x)}
\end{axis}
\end{tikzpicture}
\caption{\textbf{\textsc{AMK} self-tunes past cuBLAS on the L4 inference GPU, and the win grows with model size (batch-1 decode).} Gray dots: each config the search evaluates at $3.5$B (kernel-only, per-sample paired-interleaved, correctness-gated; ratio $=$ cuBLAS$\,/\,$\textsc{AMK} int8). Solid green: best-so-far at $3.5$B, reaching $\mathbf{1.28\times}$ (p10 $1.26$; 36/36 configs passed the dequant-reference gate), this curve is the original search run (\texttt{int8\_l4\_trajectory.json}, $n{=}80$); the higher-sample re-sweep in Table~\ref{tab:int8} reports $1.32\times$ for the same $3.5$B point ($n{=}120$). Dashed blue: best-so-far at $2.7$B, where the un-tuned default \emph{loses} ($0.97\times$) and the search crosses parity to $1.19\times$. The near-lossless (W8A16) win climbs $1.18\!\to\!1.25\!\to\!1.32\!\to\!1.33\times$ over $1.3\!\to\!2.7\!\to\!3.5\!\to\!4$B (peak $1.33\times$ at $4$B, still climbing): bigger GEMV-dominated models amortize the megakernel's fixed cross-SM sync. The same win holds across the inference fleet (L40S $1.25$--$1.27\times$, A10G up to $1.08\times$ at scale); on the high-bandwidth training-class A100/H100 cuBLAS stays ahead (Table~\ref{tab:int8}).}
\label{fig:l4-selfimprove}
\end{figure}

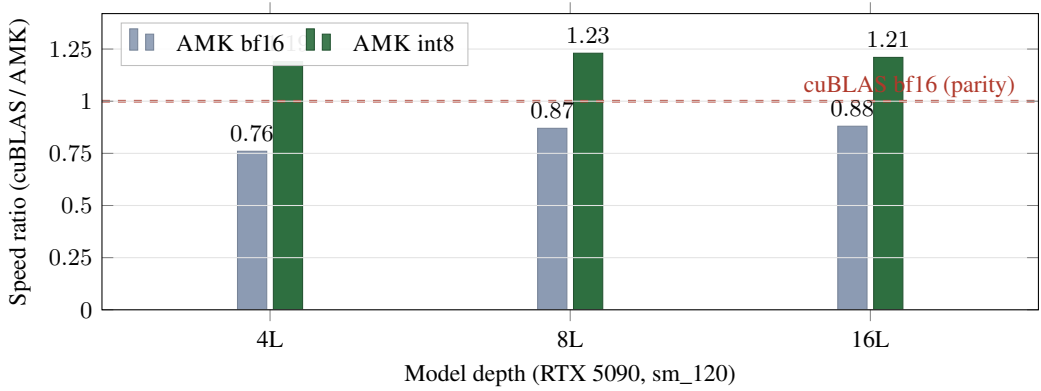
\begin{figure}[t]
\centering
\begin{tikzpicture}
  \definecolor{amkbf}{HTML}{8C9BB3}
  \definecolor{amkint}{HTML}{2E6F40}
  \definecolor{paritygray}{HTML}{B0392B}
  \begin{axis}[
    width=\linewidth,
    height=5.5cm,
    ybar=2.5pt,
    bar width=11pt,
    enlarge x limits=0.28,
    ymin=0, ymax=1.42,
    ytick={0,0.25,0.5,0.75,1.0,1.25},
    yticklabel style={font=\footnotesize},
    ylabel={Speed ratio (cuBLAS\,/\,\textsc{AMK})},
    ylabel style={font=\small},
    symbolic x coords={4L, 8L, 16L},
    xtick=data,
    xticklabel style={font=\small},
    xlabel={Model depth (RTX 5090, sm\_120)},
    xlabel style={font=\small},
    ymajorgrids=true,
    grid style={gray!22,line width=0.4pt},
    axis on top,
    legend style={
      font=\small,
      at={(0.02,0.97)}, anchor=north west,
      legend columns=2, column sep=6pt,
      draw=gray!50, fill=white, fill opacity=0.92, text opacity=1,
      /tikz/every even column/.append style={column sep=6pt},
    },
    nodes near coords,
    nodes near coords style={font=\footnotesize, /pgf/number format/fixed, /pgf/number format/precision=2},
    every node near coord/.append style={anchor=south, yshift=0.5pt},
  ]
    \addplot[draw=amkbf!85!black, fill=amkbf] coordinates {(4L,0.76) (8L,0.87) (16L,0.88)};
    \addplot[draw=amkint!75!black, fill=amkint] coordinates {(4L,1.19) (8L,1.23) (16L,1.21)};
    \draw[paritygray, dashed, line width=1pt]
      ({rel axis cs:0,0}|-{axis cs:4L,1.0}) -- ({rel axis cs:1,0}|-{axis cs:4L,1.0});
    \node[paritygray, font=\footnotesize, anchor=east] at (rel axis cs:0.985,0.755) {cuBLAS bf16 (parity)};
    \legend{\textsc{AMK} bf16, \textsc{AMK} int8}
  \end{axis}
\end{tikzpicture}
\caption{Consumer win on RTX 5090: per-token decode speedup vs.\ CUDA-graphed cuBLAS bf16 (dashed line $=1.0$ parity). \textsc{AMK} int8 weight-only (near-lossless W8A16) clears parity at every depth ($1.19$--$1.23\times$, i.e.\ $+19$--$23\%$), while \textsc{AMK} bf16 trails cuBLAS ($0.76$--$0.88\times$). This is a \emph{precision-asymmetric} win: \textsc{AMK} int8 (W8A16) vs.\ cuBLAS bf16. The bf16 bars (below parity) are the equal-precision control: per byte, \textsc{AMK}'s GEMV is still slower than cuBLAS; the int8 win comes from streaming $\approx$$0.61\times$ the weight bytes at near-lossless quality, not from a faster kernel. Ratios are cuBLAS$/$\textsc{AMK} median, batch-1, kernel-only, correctness-gated (Table~\ref{tab:int8}).}
\label{fig:consumer-win}
\end{figure}

\paragraph{Baselines, including where \textsc{AMK} loses.} Table~\ref{tab:base} is the honest comparison. All RTX 5090 rows are measured on the \emph{same} 622.9 MB 4-layer model and the same single-token decode, correctness-gated and per-sample paired-interleaved (\texttt{vs\_cublas\_10min.json}); the auto-tuned megakernel ($\approx$$1.23$ ms/token) beats naive per-op eager ($\approx$$4.44$ ms) by $3.6\times$ by removing the dozens of per-op launches and inter-op HBM round-trips that dominate eager at this scale. But CUDA-graphed eager ($\approx$$1.08$ ms) still beats \textsc{AMK} by $1.13\times$ ($\approx$$13\%$): graph replay keeps cuBLAS-quality GEMMs while amortizing launch overhead, and \textsc{AMK}'s GEMV, though much improved, remains below cuBLAS. The gap is now small but real. On the datacenter GPUs the vLLM comparison was measured on the real SmolLM2-135M checkpoint with an \emph{earlier} \textsc{AMK} GEMV (\texttt{h100.json}/\texttt{a100.json}, predating the optimized kernel above, so these rows are conservative): on H100, \textsc{AMK} ran at 14.30 ms/token versus vLLM's default cudagraph 8.65 ms/token ($1.65\times$ slower) and its \texttt{enforce\_eager} 13.51 ms/token; on A100, \textsc{AMK} 15.59 ms beat vLLM \texttt{enforce\_eager} 32.06 ms. \textbf{Disclosure:} the vLLM measurements used \texttt{dtype=float32} (not vLLM's default bf16, which would be faster), and the A100 row is \texttt{enforce\_eager} only (no CUDA-graph baseline was collected on A100). The fair default-mode comparison is the H100 cudagraph row, where \textsc{AMK} loses. We do not claim to beat vLLM.

\begin{table}[t]
\centering
\caption{Single-stream decode baselines. RTX 5090 rows: same 622.9 MB 4-layer model, single-token decode, correctness-gated, per-sample paired-interleaved (\texttt{vs\_cublas\_10min.json}, auto-tuned megakernel; CUDA-graphed cuBLAS~\cite{nvidia_cublas} as the eager baseline). Datacenter rows: real SmolLM2-135M, an \emph{earlier} \textsc{AMK} GEMV (\texttt{h100.json}/\texttt{a100.json}), so conservative. \textbf{Disclosure:} vLLM was run with \texttt{dtype=float32} (not its bf16 default, which would be faster); the A100 row is \texttt{enforce\_eager} only (no CUDA-graph A100 baseline collected). Bold marks the faster system per comparison.}
\label{tab:base}
\small
\begin{tabular}{lllr}
\toprule
GPU & Comparison & ms/token & Verdict \\
\midrule
RTX 5090 & \textsc{AMK} 1.23 vs.\ eager per-op 4.44   & 1.23 / 4.44  & \textbf{\textsc{AMK} 3.6$\times$ faster} \\
RTX 5090 & \textsc{AMK} 1.23 vs.\ CUDA-graph eager 1.08 & 1.23 / 1.08  & \textbf{cudagraph 1.13$\times$ faster} \\
H100\textsuperscript{$\dagger$}     & \textsc{AMK} 14.30 vs.\ vLLM cudagraph 8.65 & 14.30 / 8.65 & \textbf{vLLM 1.65$\times$ faster} \\
H100\textsuperscript{$\dagger$}     & \textsc{AMK} 14.30 vs.\ vLLM enforce-eager 13.51 & 14.30 / 13.51 & \textbf{vLLM 1.06$\times$ faster} \\
A100\textsuperscript{$\dagger$}     & \textsc{AMK} 15.59 vs.\ vLLM enforce-eager 32.06 & 15.59 / 32.06 & \textbf{\textsc{AMK} 2.06$\times$ faster} \\
\bottomrule
\end{tabular}

\vspace{2pt}
{\footnotesize \textsuperscript{$\dagger$}Datacenter rows predate the optimized GEMV (earlier \textsc{AMK}); re-measurement is pending.}
\end{table}

The picture is consistent. \textsc{AMK} wins where the competition is per-op launch overhead and loses where the competition is kernel quality. The launch-fusion win is real and additive with future kernel-efficiency work, but it does not yet close the gap to graph-replayed cuBLAS or to vLLM's default path.

\section{Discussion and Limitations}
\label{sec:limitations}

\textbf{Kernel quality below cudagraph and vLLM.} On the optimized bf16 GEMV \textsc{AMK} reaches $\approx$$460$ GB/s on the 622.9 MB model ($\approx$$51\%$ of spec / $\approx$$63\%$ of measured peak), versus a cuBLAS bf16 ceiling of $\approx$$661$ GB/s ($\approx$$90\%$ of measured peak), and at the whole-decode level loses to CUDA-graphed cuBLAS by $1.13\times$ ($\approx$$13\%$); on the datacenter GPUs, with an earlier GEMV, it reached 1--16\% of spec (12.5--17.7\% of measured A100 peak, 4.8--8.6\% of measured H100 peak, clock-pinned) and lost to default vLLM by $1.65\times$. The diagnosed cause is the v1 GEMV: a coalesced-scalar warp-per-row dot product with a serial $K$-loop, no tensor-core path (which a memory-bound batch-1 GEMV would not benefit from anyway), and a grid-wide counter sync per tile. The next lever is more memory-level parallelism (deeper load pipelining, \texttt{cp.async} multi-buffered SMEM staging) and far fewer, coarser sync points. The correctness-bearing architecture that makes such a rewrite safe and automatic is already in place; this is a kernel-quality push, not a redesign.

\paragraph{Limitations: where the auto-generated kernel does not win.} The win over cuBLAS holds across the datacenter inference-class GPUs (L4 up to $1.33\times$, the current-gen L40S $1.25$--$1.27\times$, and the A10G up to $1.08\times$ at scale) and the consumer RTX~5090 ($1.19$--$1.23\times$), but \emph{not} on the high-bandwidth training-class A100/H100 ($1.4$--$3.0$ TB/s), where the auto-generated int8 megakernel \emph{trails} cuBLAS. Self-tuning the int8 knobs on each part (not transplanting the laptop config), the best per-arch config narrows the gap monotonically with scale but plateaus below parity: A100 reaches ${\approx}0.79\times$ at $1.3$B down to ${\approx}0.55\times$ at $13$B, and H100 ${\approx}0.72\times$ down to ${\approx}0.60\times$. We treat this as a result, not an apology: the harness \emph{measured} the boundary and then \emph{localized its cause}, by directly A/B-testing the two obvious levers: both regressed. A \texttt{cp.async} weight-staging ring for the int8 GEMV (the same double-buffered SMEM ring \textsc{AMK}'s fp GEMV already uses, mirrored to stage raw int8 granules and dequantize per group) was \emph{slower} than the synchronous path (ring$/$sync $0.82\times$ on the A100, $0.87\times$ on the L4), so the decode GEMV is not load-latency-bound; and split-KV attention, which likewise adds cross-SM synchronization, also regressed. Together these isolate the \emph{per-tile cross-SM synchronization} (a fixed cost the megakernel pays and cuBLAS does not, worst exactly on the fastest training silicon) as the structural binder; the remaining lever is a coarser-sync scheduler (fewer grid-wide barriers per layer), which is future work, not GEMV load-pipelining or more aggressive quantization. The ordering is \emph{not} a clean function of bandwidth: the $864$\,GB/s L40S wins by more than the $600$\,GB/s A10G, the deficit is worst on the highest-bandwidth A100/H100, and the Turing T4 stays just below parity ($0.95$--$0.97\times$) for the orthogonal reason of occupancy ($64$\,KB SMEM $\Rightarrow$ one block/SM) despite low bandwidth, and so does not cross at batch-1 pos-0 either. This is the honest scope of single-stream megakernels: they help on inference-class GPUs at batch~1, and a coarser-sync GEMV is what would extend the win onto training-class silicon.

\textbf{Clocks and the spec-versus-measured denominator.} The laptop GPU is power-capped (it begins a run at 180 MHz and ramps under load) and the datacenter SM clocks were not pinned in the main study, which inflates both latency and variance. The clock-pinned re-measurement (Table~\ref{tab:pinned}) shows the roofline fraction is within $\pm0.8$ percentage points of the unpinned numbers, confirming the gap is kernel quality, not throttling. We report both the spec and the measured HBM peak as denominators; measured peak is the fairer floor but does not close the gap.

\textbf{Quantization is modest, and int4 is lossy.} The auto-generated int8 megakernel is greedy-lossless but only $1.12\times$ faster per token; the int4 path drops the weight-traffic floor $2.42\times$ but with naive round-to-nearest is lossy ($\approx$22\% greedy-token agreement). Dequant ALU and Amdahl's law on the non-GEMV work cap the speedup; a calibrated int4 scheme and a quantization-aware GEMV are future work.

\textbf{No hardware counters.} ncu/Nsight was unavailable on our Modal account, so every utilization number is wall-clock plus analytic roofline, not a measured DRAM-throughput counter. Pinning clocks and capturing ncu traces are the obvious next measurement steps.

\textbf{Position-0, empty-KV measurement.} Every reported decode latency is a single batch-1 step at position~0 with an empty KV cache; it is weight-dominated by construction and matches the bandwidth-bound floor we study, but it does not capture the attention/KV-read cost that grows with context length. Long-context per-token latency, where KV traffic becomes a second bandwidth term, is not measured here and is future work.

\textbf{A self-funded effort with constrained compute.} This research was self-funded by the authors, with no institutional or industry compute grant; every datacenter measurement was a short, budgeted cloud rental (on the order of a few GPU-hours in total). That budget bounded the experimental scope: we could not run the long, sustained autotuning/search campaigns, nor the iterative tensor-core/DP4A GEMV development and on-hardware tuning, that would be required to push \textsc{AMK}'s kernel quality toward hand-built, manually-optimized novel megakernels on datacenter silicon. The datacenter performance gap therefore reflects, in part, optimization effort and compute we could not afford rather than a fundamental ceiling; the correctness-bearing architecture is already in place to absorb that work once resources allow. We report only what we could afford to measure, and we report it plainly.

\textbf{Architecture coverage.} Self-retargeting is proven on the bias-free, full-rotary, SiLU-SwiGLU, RMSNorm, GQA Llama family across sm\_80/90/120 (10 of 10 supported models). Other HF architectures (MoE routing, sliding-window or fused-QKV attention, partial or scaled RoPE, biased projections) are out of scope and rejected at import, with one documented exception: a Qwen2 with hardcoded q/k/v biases passes the config-only check and is silently accepted, which a state-dict bias scan would close. bf16 logit error on a real checkpoint exceeds the strict fp32 tolerance (we gate bf16 on token agreement, which holds). The \emph{harness} itself, however, is not Llama-specific: the schedule-IR validator, the reference-oracle correctness gate, and the propose/evaluate/keep loop are architecture-, language-, and target-agnostic. Generalizing the importer (other architectures), the backend (beyond CUDA), and the deployment targets is future work that this agent-driven loop is built to absorb rather than a redesign; the agent is what supplies that generality, and broadening coverage into a general megakernel-synthesis harness is the central direction of the work going forward.

\section{Conclusion}

\textsc{AMK} compiles a HuggingFace Llama-family model into a single persistent cooperative kernel with no per-model hand CUDA, and makes safety a static property: across 7{,}160 adversarial schedules the validator had zero false-accepts, so an unsafe agent-proposed schedule is rejected rather than hung. It auto-generates a correct megakernel for all 10 supported models (including three real checkpoints up to TinyLlama-1.1B) and for int8 and int4 weight-only quantized variants, the int8 path lossless. The same source retargets across sm\_80, sm\_90, and sm\_120 with the gencode derived from the live device, and on a real SmolLM2-135M checkpoint reproduces HuggingFace greedy decode token-for-token and matches its perplexity to $2.5\times10^{-7}$. The performance study is deliberately honest and reports both directions: a search-found int8 weight-only megakernel \emph{beats} CUDA-graphed cuBLAS bf16 at batch-1 across the datacenter inference fleet: the L4 by up to $1.33\times$, the current-gen L40S by $1.25$--$1.27\times$, and the A10G by up to $1.08\times$ at scale (a precision-asymmetric W8A16-vs-bf16 win that holds across the inference-class regime but not the training-class A100/H100, and is not a clean function of bandwidth since the $864$\,GB/s L40S wins by more than the $600$\,GB/s A10G), while on the equal-precision bf16 path it trails CUDA-graphed cuBLAS by $1.13\times$ ($\approx$$13\%$) and default vLLM by $1.65\times$, its optimized GEMV reaching $\approx$$63\%$ of measured HBM peak versus a cuBLAS ceiling of $\approx$$90\%$; the knob-autotuning loop finds these operating points automatically. The path to closing the equal-precision gap is a higher-bandwidth GEMV with coarser synchronization, and the statically-checked compiler that makes that work safe, and that self-improvement automatic, is the contribution we offer now.

\end{document}